# Synthetic Database for Evaluation of General, Fundamental Biometric Principles

Lee Friedman and Oleg Komogortsev
Department of Computer Science
Texas State University, San Marcos, TX

## Abstract

*We create synthetic biometric databases to study general, fundamental, biometric principles. First, we check the validity of the synthetic database design by comparing it to real data in terms of biometric performance. The real data used for this validity check was from an eye-movement related biometric database. Next, we employ our database to evaluate the impact of variations of temporal persistence of features on biometric performance. We index temporal persistence with the intraclass correlation coefficient (ICC). We find that variations in temporal persistence are extremely highly correlated with variations in biometric performance. Finally, we use our synthetic database strategy to determine how many features are required to achieve particular levels of performance as the number of subjects in the database increases from 100 to 10,000. An important finding is that the number of features required to achieve various EER values (2%, 0.3%, 0.15%) is essentially constant in the database sizes that we studied. We hypothesize that the insights obtained from our study would be applicable to many biometric modalities where extracted feature properties resemble the properties of the synthetic features we discuss in this work.*

## 1. Introduction

Biometric researchers have been employing synthetic biometric databases for several years [1-8]. Typically, the effort is to create synthetic images (Face [4], Iris [3], Fingerprint [2, 8]). These are used to address practical issues such as removing the privacy concerns that surround real data [1, 4], creating very large datasets without collecting data from many human subjects [3], training purposes [2], and to study forgery strategies [7].

In contrast, in the present study, when we refer to synthetic databases, we are talking about creating spreadsheets of data where the 1st column presents the subject ID, the 2nd column represents the recording session ("visit" or "occasion" when biometric template was taken), and the remaining columns represent synthetically generated features. When we create these databases, we control the number of features that are included, the degree of variation in the temporal persistence of the features, the number of subjects in the database, and, potentially, the degree of intercorrelation among features. The goal of creating such databases is to study general, fundamental biometric principles given certain assumptions on the feature data.

The first goal of our work is to study the impact of variations in temporal persistence on biometric performance. The temporal persistence is understood as the tendency of a feature to remain stable over the period of time between gallery and probe. While the term "permanence" is commonly used by the biometrics community, biometric features do not actually need to be permanent, just temporally persistent over the relevant test-retest time intervals, which in some cases are minutes, days, or months.

Prior research has shown [9] that the intraclass correlation coefficient (ICC) [10, 11] can index the temporal persistence of a single biometric feature, and that the degree of temporal persistence of a set of features has a substantial impact on the biometric performance. In this paper, we study this phenomenon further by creating synthetic features with various simulated levels of temporal persistence.

A second goal of our work is to study the impact of the number of subjects in a database on biometric performance. We create synthetic databases, with from 100 to 10,000 subjects and study biometrics performance. We hypothesized that the number of features required to achieve a particular biometric performance (e.g., EER <= 2%) should increase as the number of subjects in the biometric database increases. In this work we test this hypothesis.

Our report is organized as follows: section 2 presents the process of feature creation, section 3 defines the measurement of temporal persistence, section 4 discusses the creation of feature sets, section 5 presents biometric performance results for synthetic and real features, biometrics results obtained from the synthetic features from various temporal persistence groups, and the impact of the number of subjects on the resulting biometrics performance, Section 6 presents a discussion and conclusion.

## 2. Synthetic Features & their Characteristics

Let us define $X_i^j$ as a feature captured during a recording session (visit or occasion) *j,* where *j* =[1..*k*], from subject *i,* where *i*=[1..*n*]. In this work, we only use two recording sessions, thus *k*=2. Let us use the notation "RSND(0,1)" to mean a function which returns a standard normal deviate, which means a random value from a normal distribution with mean = 0 and standard deviation (SD) = 1.0.

To generate a feature $X_i^j$ for recording sessions 1 and 2, the following steps are taken:

$$X_i^1 = RSND(0,1) \quad 1$$

$$X_i^2 = X_i^1 \quad 2$$

$$X_i^1 = X_i^1 + RSND(0,1) * Mult \quad 3$$
$$X_i^2 = X_i^2 + RSND(0,1) * Mult$$

where *Mult* is a multiplier. These calcutions are performed across all subjects i (i=1…$n$). The larger the multiplier the less similar the two recording sessions are. The utility of the multiplier is to control the degree of the temporal persistence and error that we discuss in the next subsection. Section 4 describes ranges of multipliers that allow us to produce targeted levels of temporal persistence. These features are generally normally distributed. As a final step, the values for each feature are z-score transformed with mean = 0 and SD = 1.0).

We will be comparing our synthetic features to real eye movement-derived biometric features captured from two recording sessions, which have also been z-score-transformed (mean = 0, SD = 1.0). This standardization of the variance of each feature puts each feature on an equal footing. Otherwise, features with greater variance will contribute more strongly to the subsequent N-dimensional distance calculations, allowing such variables to dominate biometric performance.

## 3. Measurement of Temporal Persistence

We assess temporal persistence using the ICC [10, 11]. It is somewhat similar to the Pearson r correlation coefficient, but it is designed not to evaluate the relationship between two classes of measurement (e.g., weight (kg) versus height (cm)), but rather to evaluate the relationship between several measurements of the same class, such as the measurement of the same biometric feature from the same set of subjects, during repeated recording sessions. There are several types of ICC and several ways to calculate it [10, 11]. Below, we present one method to calculate the ICC. Prior to using the ICC, it is suggested that the reader review two seminal references [10, 11].

Under the assumption of the two-way random effects model, each feature $X_i^j$ can be represented in the following way:

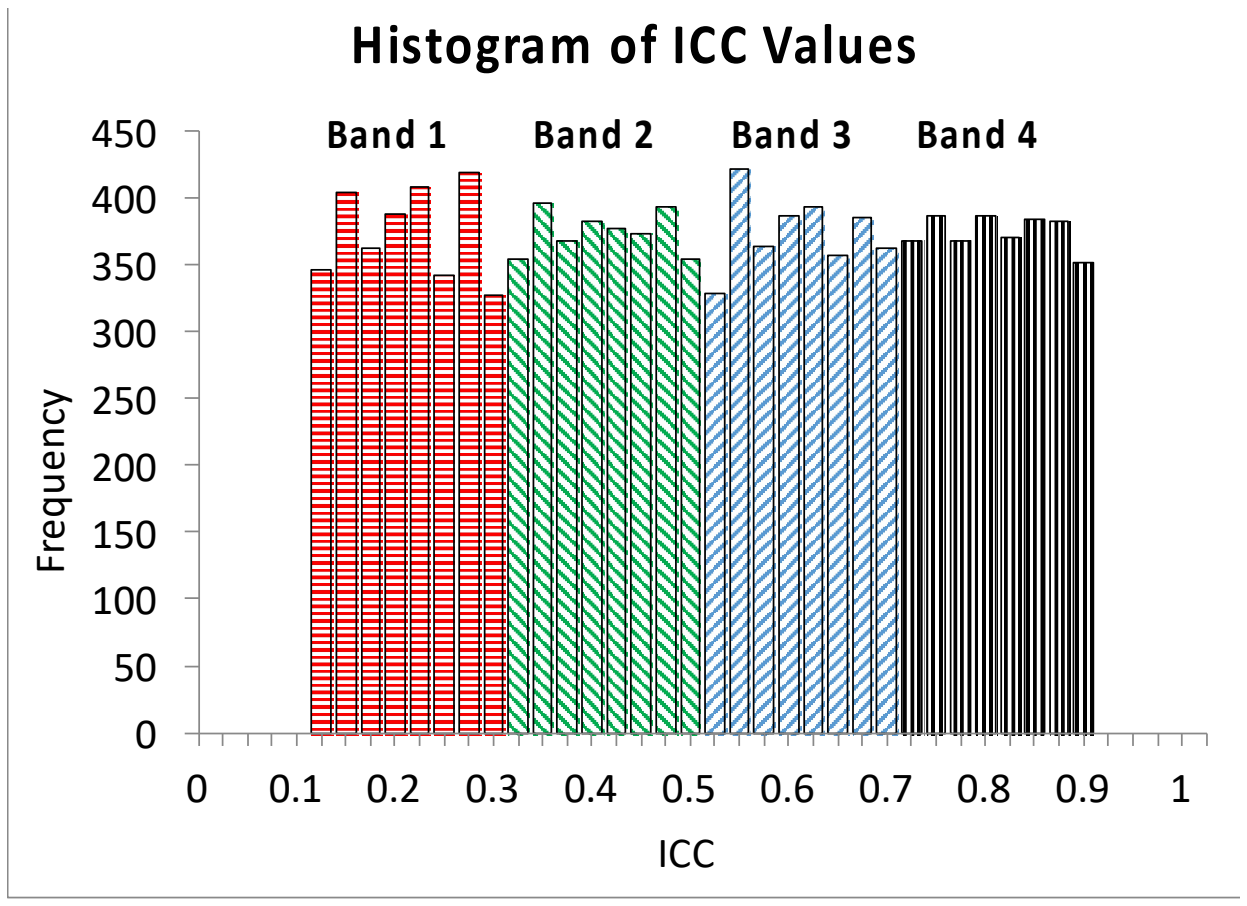

**Figure 1.** Histogram of ICC values for different ICC bands.

$$X_i^j = S_E_i + O_E_j + Er_{ij} \qquad 4$$

Where $S_E_i$ is the subject effect, which is true unobservable level of the feature *m* for subject *i*. $S_E_i$ *is* assumed to be normally distributed with mean = 0 and variance = $\sigma_s^2$. $O_E_j$ is the occasion effect which is the random effect on the feature of the time between recording occasions $j$. $O_E_j$ *is* assumed to be normally distributed with mean = 0 and variance = $\sigma_o^2$. $Er_{ij}$ is random error, that is not accounted for by subject and occasion effects, and consists of many, presumably small random unknown factors. For example, measurement error would contribute to random error. $Er_{ij}$ is also assumed to be normally distributed with mean = 0 and variance = $\sigma_e^2$. Please note that values $\sigma_s^2$, $\sigma_o^2$, and $\sigma_e^2$ are feature specific. Under this model, we assume that all effects are mutually independent. With all these assumptions the intraclass correlation coefficient (ICC) for the feature $X_i^j$ can be computed as:

$$ICC = \frac{Subject_Variance}{Total_Variance} = \frac{n \cdot (MS_S_E - MS_Er)}{n \cdot MS_S_E + k \cdot MS_O_E + (n \cdot k - n - k) \cdot MS_Er} \qquad 5$$

where prefix MS stands for mean square. All corresponding mean squares for $S_E$, $O_E$, and $Er$ are obtained from a fixed-effect two-way ANOVA [10] as applied to the generated features created by equations (1)-(3).

The ICC ranges from 0.0 to 1.0. In general use, the ICC is conceived of as an index of the reliability of a measurement. Cichetti, and Sparrow [12, 13] have suggested that ICC levels be interpreted by the following rules of thumb: ICC >= 0.75: "Excellent Reliability", ICC >= 0.60, and < 0.75: "Good Reliability", ICC >= 0.40, and < 0.60: "Fair Reliability", and ICC < 0.40: "Poor Reliability".

## 4. Creation of Feature Sets that Meet Specific Temporal Persistence Levels

The value of the *Mult* variable from equation (3) determines the similarity of the data recorded from session 1 to session 2. We vary the level of the *Mult* variable randomly from 1.4 to 2.8 to produce features with ICCs from 0.1 to 0.3; from 0.9 to 1.7 to produce features with ICCs from 0.3 to 0.5; from 0.6 to 1.0 to produce features with ICCs from 0.5 to 0.7; and from 0.3 to 0.7 to produce features with ICCs from 0.7 to 0.9. The step size between each range of *Mult* was 0.01. We refer to 4 ICC bands: Band 1 where ICC varies in the interval [0.1 to 0.3], Band 2 [0.3 to 0.5], Band 3 [0.5 to 0.7] and Band 4 [0.7 to 0.9]. Within each band, as we create new features, we randomly select the value of the *Mult* variable from the ranges above. We keep a running total of the number of features created in each band in order to create an even ICC distribution, with equal numbers of features in each band. See Figure 1 for a frequency histogram of 12,000 synthetic features (3000 features for each ICC Band) created as above.

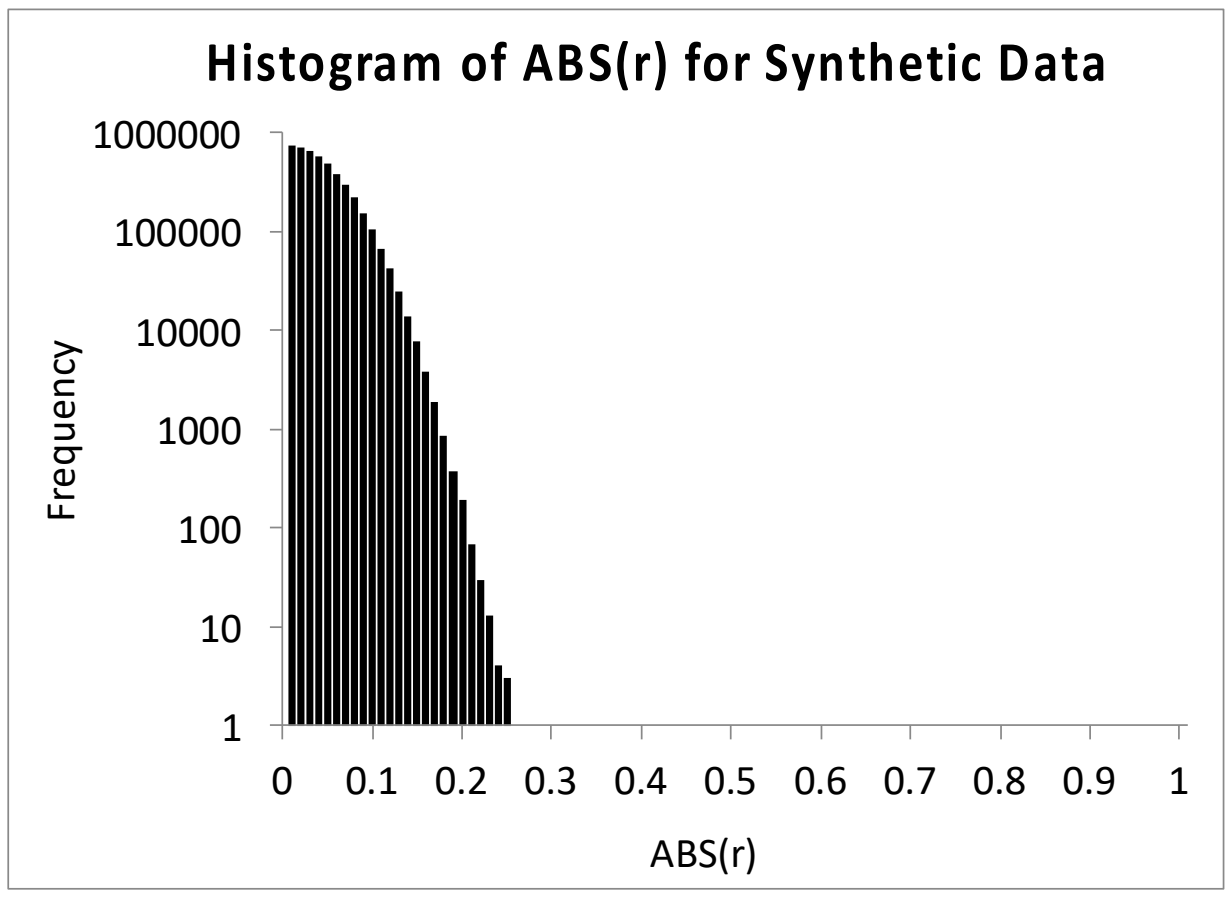

**Figure 2.** Histogram of synthetic intercorrelations.

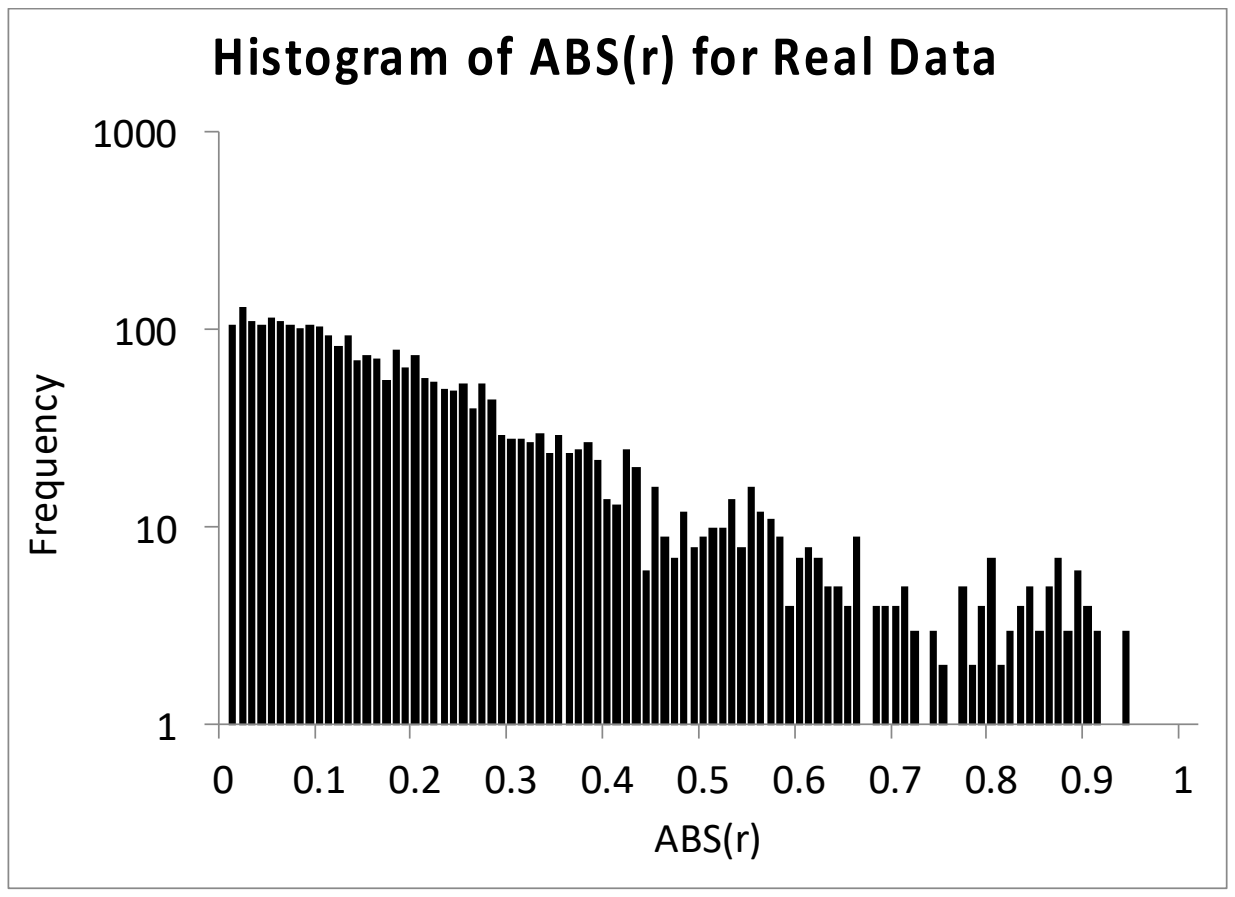

**Figure 3.** Histogram of real data intercorrelations.

## 5. Results

Comparative Biometric Performance of Real and Synthetic Features

To validate our synthetic database, we compared the biometric performance of our synthetic features to real features from a large set of eye-movement related features [9]. Specifically, we employed the features from the "EM-2-ST" dataset from [9], with more than 300 features. These real eye movement features were either normally distributed or were transformed to normality using classical data transformations [9, 14, 15]. They were also z-scored transformed (mean 0, S.D. = 1.0).

From these features, we selected those which had ICC values in the same ranges as our Band 1 to Band 4 synthetic datasets. We next compared the intercorrelations between our synthetic features to the intercorrelations between our real data features.

For example, let us compare the absolute value of the intercorrelations between synthetic features, on the one hand, and real data features, on the other hand, for ICC Band 3 (ICC: 0.5 to 0.7). (Note, the same pattern holds for all bands.) In Figures 2 and 3, we present frequency histograms of the absolute value of all intercorrelations between synthetic features and eye movement-derived features respectively.

The median Pearson r (absolute value ABS(r)) correlation coefficient for the synthetic features was 0.03 ($95^{th}$ percentile = 0.09) and for the real features was 0.14 ($95^{th}$ percentile = 0.58). Obviously, real data are much more intercorrelated than synthetic data.

In the future, we plan to create and study synthetic datasets that mimic the intercorrelation structure of real data, but for the present, we find it useful to compare the biometric performance of the current synthetic features with real data. To minimize the differences between real and synthetic data *a priori*, we chose the largest subset of real data features that had, at most, a maximum intercorrelation of 0.3. For Band 3, we found 19 such features. Please note that even with this restriction, real data are still more intercorrelated than synthetic data.

In Figure 4, we compare the equal error rate (EER)[1] of real data features (black filled dots) with that of synthetic features (gray filled dots) for ICC Band 3, as the number of features used for biometric verification increases from 2 to 19 features. Each dot in Figure 3 represents the median of 10 datasets, each with a randomly selected subset of features. Note the highly

[1] Please note that to compute the EER we follow the same methodology of computation, cross validation and distance as described in reference [9].

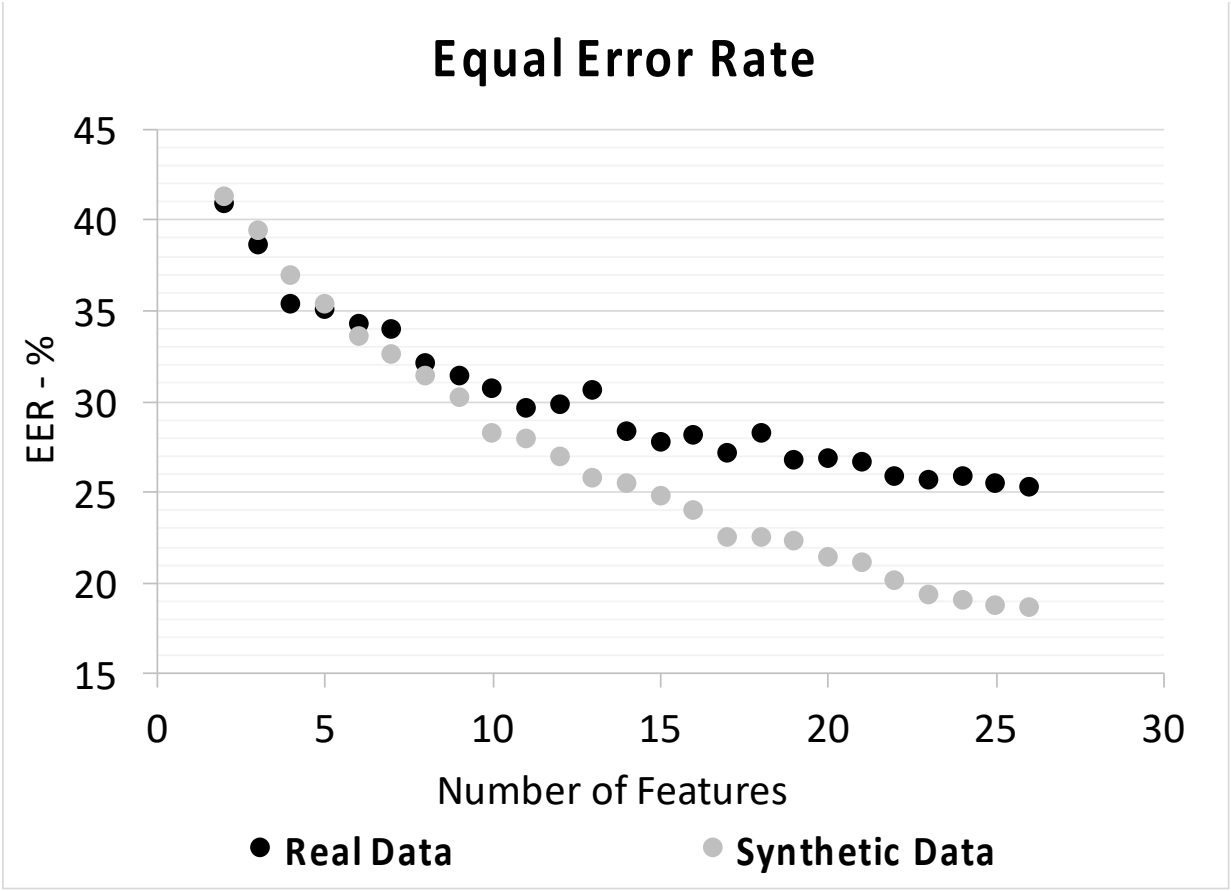

**Figure 4.** Real vs synthetic biometric performance.

similar EER values for real data and synthetic data for from 2 to 6 features.

As we move from 7 to 26 features, the synthetic features perform better than the real features, with an increasing advantage to synthetic features as the number of features used approaches 26. We hypothesize that the improved biometric performance of the synthetic features is due to the lower intercorrelation of the synthetic features – in this case, each synthetic feature contains more unique identifying information than each real feature.

Once we create synthetic datasets with intercorrelations that mimic real data we will be able to test this hypothesis. We conclude that, but for the level of intercorrelation, the biometric performance resulting from the synthetic features is quite similar to the biometric performance coming from real eye movement features.

Comparative Biometric Performance of our 4 ICC Bands

We assessed the biometric performance of ICC Bands of synthetic features. We started with 4 synthetic databases, one for each ICC Band, and each modelled as having 3000 features and 500 subjects tested on 2 occasions. We assessed biometric performance with the EER. We included from 2 up to 100 features chosen randomly. The results are presented in Figure 5. Each marker on this plot is the median of 25 randomly chosen feature datasets. It is clearly visible that the high ICC group (Band 4, ICC: 0.7 to 0.9), performs much better that the other bands, achieving perfect performance (EER = 0.0) with 35 features. Band 3 performs better than Band 2 and Band 2 performs better than Band 1. This plot illustrates the substantial impact of variations in temporal persistence on biometric performance.

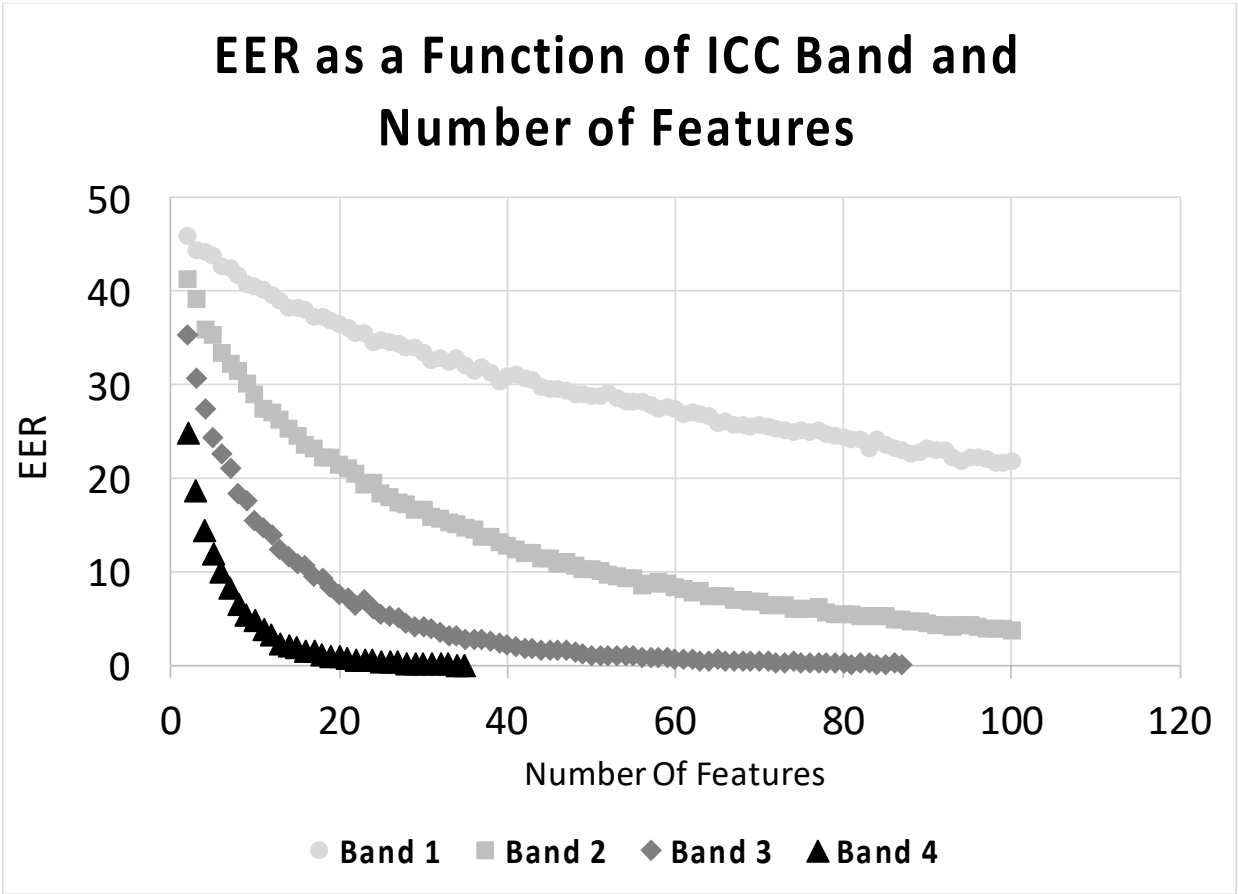

**Figure 5.** Relationship between number of features and EER.

We wanted to know how the change from one ICC Band to another affected the similarity scores for genuine and impostor distributions. We compared the similarity scores for 25 sample datasets per ICC Band, based on the inclusion of 25 randomly chosen features. Our results are illustrated in Figure 6. The black squares represent the median values of the genuine similarity score distributions across ICC Bands, and the open black circles represent the median values of the impostor distributions. Note the substantial increase in the median of the genuine similarity score distributions as we go from ICC Band 1 to ICC Band 4. In contrast, there was no association for the medians of the impostor distributions. In Figure 7, we present the interquartile range (IQR) for the genuine and impostor similarity score distributions as a function of ICC Band. Note the impressive decrease in the IQR of the genuine distributions, and the absence of significant change for the IQR of the impostor distributions. Thus, we can conclude that higher ICC bands improve biometric performance by increasing the median, and decreasing the spread of the genuine similarity score distributions. Obviously, such changes will lead to superior performance of high ICC bands.

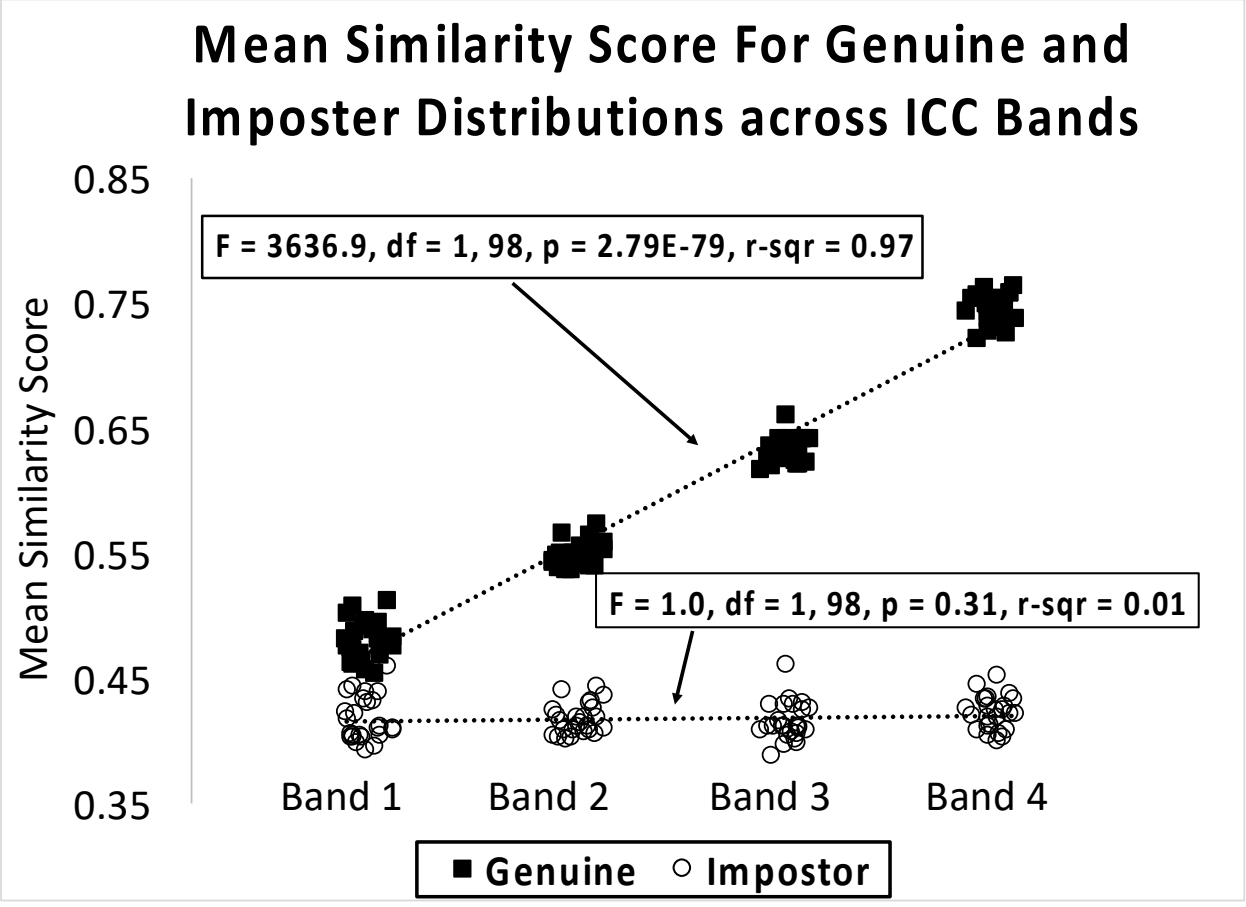

**Figure 6.** Relationship between median similarity score and ICC Band.

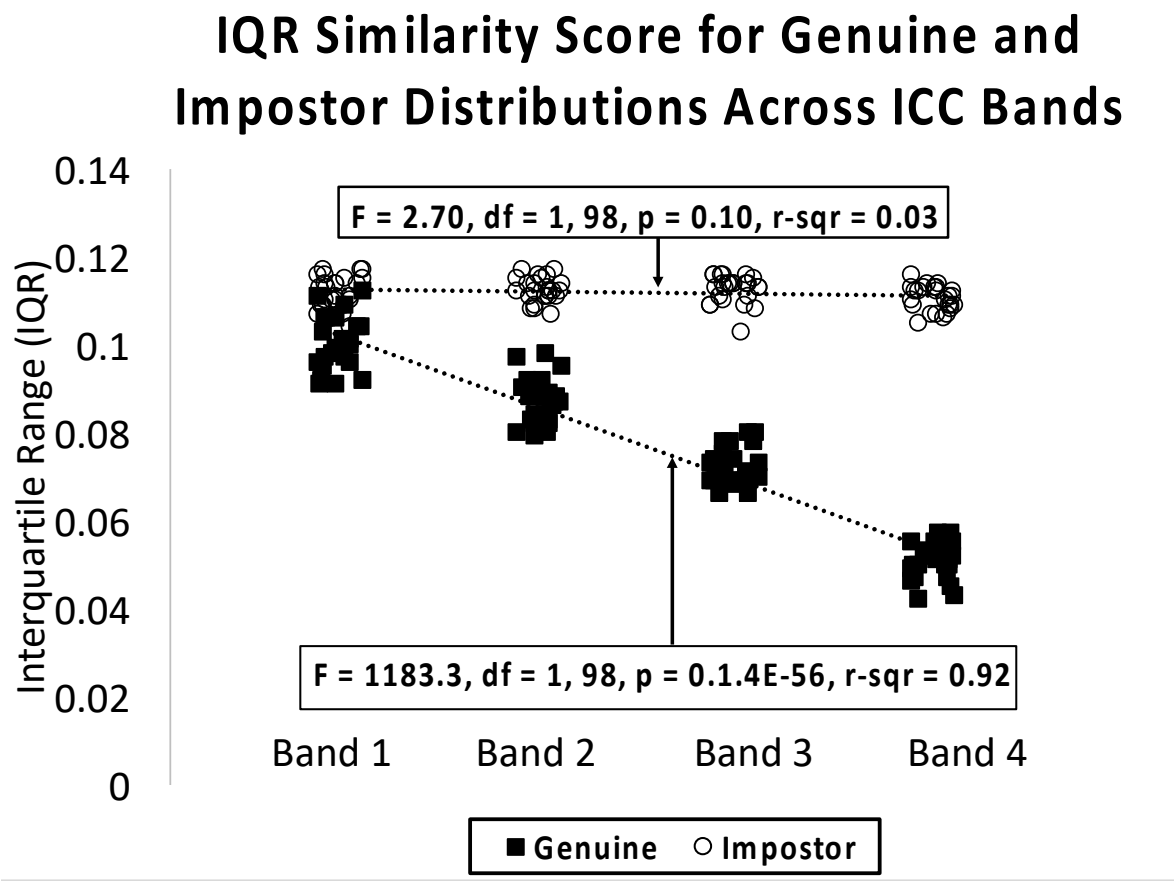

**Figure 7.** Relationship between IQR similarity score and ICC band.

Impact of the Number of Subjects on Biometric Performance

We hypothesized that as the number of subjects in a database increased from 100 to 10,000, more and more features would be required to achieve a particular level of performance (e.g., EER <= 2%). For this analysis, we studied only Band 4

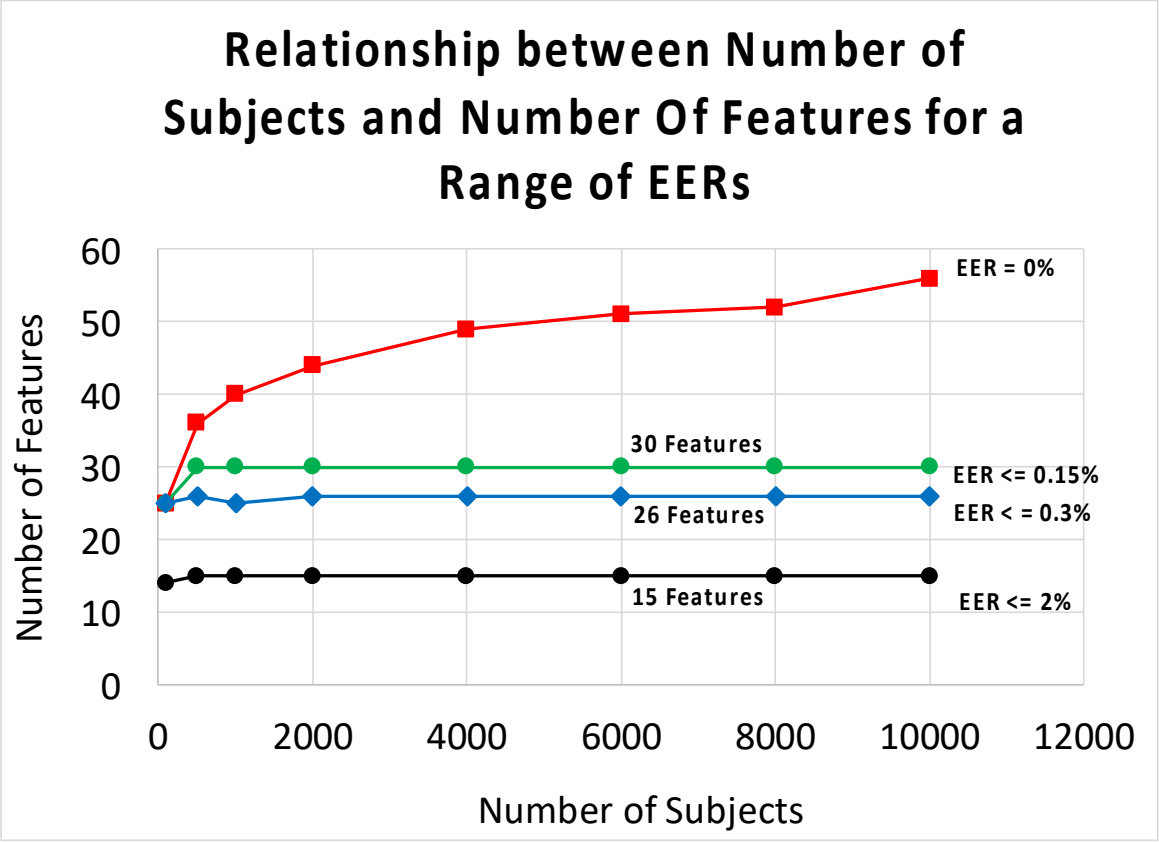

**Figure 8.** Relationship between number of subjects and number of features required to achieve several levels of performance

datasets (ICC: 0.7 to 0.9). We created synthetic databases with *n* = 100, 500, 1000, 2000, 4000, 6000, 8000 and 10000 subjects. We determined the number of features required to achieve several levels of biometric performance (EER <= 2%, EER <= 0.3%, EER <= 0.15% and EER = 0%). The results are presented in Figure 8. Each marker in this plot is the median of 100 datasets with randomly chosen features from the overall database. We found that for several low EER performance benchmarks (EER <= 2%, EER <= 0.3% and EER <= 0.15%) the number of features remained constant (15, 26 and 36 features respectively) as the number of subjects increased from 500 subjects to 10,000 subjects. However, if the goal was to achieve perfect performance (EER = 0%) the number of features increased monotonically from 25 features for 100 subjects to 55 features for 10,000 subjects.

## 6. Discussion and Conclusion

We have shown that synthetic biometric databases can be constructed that perform similarly to real data. We have used such databases to study the impact of temporal persistence as indexed by the ICC. And we have used such databases to evaluate the relationship between sample size (number of subjects in a dataset) and the number of features required to achieve particular levels of biometric performance.

The method of constructing synthetic databases that we present here produces sets of features which are essentially uncorrelated. The only correlations that occur, occur by chance. Real data is much more highly intercorrelated. One of the goals of our future work is to develop synthetic databases that mimic real data in this way. Nonetheless, we have shown that real features, with minimal intercorrelations from an eye movement dataset, perform similarly, biometrically, with synthetic data.

Our analysis of the impact of temporal persistence of synthetic biometric features, as assessed with the ICC, is very similar to the impact of temporal persistence of real features as described in [9]. The impact is substantial. Higher ICC features perform much, much better than lower ICC features. We have shown that as one moves from ICC Band 1 to ICC Band 4, the medians of the genuine similarity score distributions increase, and the spread (IQR) of the genuine similarity score distributions decrease, leading to markedly improved performance.

We had predicted that, as the number of subjects in a biometric database increased from 100 to 10,000, the number of features needed to achieve particular levels of biometric performance would increase. This hypothesis was not supported. The number of features required to achieve several low EER values did not change with increasing sample size, except for EER = 0%, which is an unrealistic goal for any real biometric authentication system.

We plan to employ our synthetic database approach in the future to ask additional questions. We hope to be able to impart varying levels of intercorrelations among synthetic features, and evaluate biometric performance as a function of intercorrelation. We also plan to compare the performance of PCA components to raw features in terms of biometric performance. We may also compare databases of normally distributed features to non-normally distributed databases, to evaluate the sensitivity of our analysis to normality and to assist in the development of methods to combine information from normally distributed features and non-normally distributed features.

In conclusion, we believe that we have presented a method to create synthetic biometric databases, which are reasonable facsimiles of real data, that can be used to address several fundamental, conceptual issues in the field of biometrics.

## Acknowledgement

We acknowledge support from the National Science Foundation for a NSF CAREER awared (#CNS-1250718) to Dr. Komogortsev.

We acknowledge support from the Nation Institute of Standards and Technoology (NIST; #60NANB15D325) to Dr. Komogortsev..

We acknowledge support from Google Inc, for a Virtual Reality Award (2016) to Dr. Komogortsev